\documentclass[letterpaper, 10 pt, conference]{ieeeconf}  

\IEEEoverridecommandlockouts                              

\usepackage{graphics} 
\usepackage[pdftex]{graphicx}
\usepackage{circuitikz}
\usepackage{tikz}
\UseRawInputEncoding
\usetikzlibrary{calc,patterns,decorations.pathmorphing,decorations.markings,decorations.pathreplacing}
\usepackage{tikzscale}
\usetikzlibrary{patterns,decorations.pathmorphing,positioning}
\usetikzlibrary{arrows}
\usepackage{scalefnt}
\usepackage{subfigure}
\graphicspath{ {./images/} }
\usepackage{amsmath}
\usepackage{float}
\usepackage{circuitikz}
\usepackage{pgfplots}
\usepgfplotslibrary{units}
\usepackage{filecontents}
\usepackage{tikz,filecontents,amsmath}
\usepackage{algorithmic}
\usepackage{amssymb}    

\usepackage{bm,times}
\newcommand{\vc}[1]{\mathbf{\bm{#1}}} 
\usepackage{bm} 
\usepackage{color} 
\usepackage{hyperref}
\usepackage{cancel}
\usepackage[export]{adjustbox}
\usepackage{mathtools}
\usepackage{caption}
\usepackage{comment}
\usepackage{fixltx2e}
\usepackage[capitalise]{cleveref}
\usetikzlibrary{calc}

\usepackage{todonotes}
\usepackage{hyperref}
\usepackage{pdfpages}
\usepackage{graphicx}
\usepackage{caption}
\usepackage{booktabs} 
\usepackage{amsmath,bm,mathtools,upgreek}
\usepackage{amssymb,stmaryrd}
\usepackage{prettyref}
\newrefformat{fig}{Figure~[\ref{#1}]}
\newsavebox\IBoxA \newsavebox\IBoxB \newlength\IHeight
\newcommand\TwoFig[6]{
  \sbox\IBoxA{\includegraphics[width=0.3\textwidth]{#1}}
  \sbox\IBoxB{\includegraphics[width=0.3\textwidth]{#4}}%
  \ifdim\ht\IBoxA>\ht\IBoxB
    \setlength\IHeight{\ht\IBoxB}%
  \else\setlength\IHeight{\ht\IBoxA}\fi
  \begin{figure}[!htb]
  \minipage[t]{0.3\textwidth}\centering
  \includegraphics[height=\IHeight]{#1}
  \caption{#2}\label{#3}
  \endminipage\hfill
  \minipage[t]{0.3\textwidth}\centering
  \includegraphics[height=\IHeight]{#4}
  \caption{#5}\label{#6}
  \endminipage 
  \end{figure}%
}
\makeatletter 
\newcommand*{\b@xplus}[1][+]{\ooalign{%
    $\m@th\vcenter{\hbox{$\m@th#1$}}$\cr%
    \hidewidth$\m@th\boxempty$\hidewidth\cr}} 
\renewcommand*{\boxplus}{\mathbin{\b@xplus}} 
\renewcommand*{\boxminus}{\mathbin{\b@xplus[-]}} 
\makeatother

\title{\LARGE \bf
A Whole-Body Controller Based on a Simplified Template for Rendering Impedances in Quadruped Manipulators
}

\author{Mattia Risiglione, Victor Barasuol, Darwin G. Caldwell and Claudio Semini
\thanks{$^{1}$Istituto Italiano di Tecnologia (IIT), Via S. Quirico 19D, 16163 Genoa (Italy).  \textit{email}: \tt\small firstname.surname@iit.it.} 
}

\begin{document}

\maketitle
\thispagestyle{empty}
\pagestyle{empty}


\begin{abstract}
Quadrupedal manipulators require to be compliant
when dealing with external forces during autonomous manipulation,
tele-operation or physical human-robot interaction. 
This paper presents a whole-body controller that allows
for the implementation of a Cartesian impedance control
to coordinate tracking performance and desired compliance
for the robot base and manipulator arm. The controller is
formulated through an optimization problem using Quadratic
Programming (QP) to impose a desired behavior for the system
while satisfying friction cone constraints,
unilateral force constraints, joint and torque limits. The presented
strategy decouples the arm and the base of the platform, enforcing
the behavior of a linear double-mass spring damper system, and
allows to independently tune their inertia, stiffness and damping
properties. The control architecture is validated through an extensive
simulation study using the 90kg HyQ robot equipped with a 7-DoF manipulator
arm. Simulation results show the impedance rendering performance
when external forces are applied at the arm's end-effector. The paper
presents results for full stance condition (all legs on the ground)
and, for the first time, also shows how the impedance rendering is
affected by the contact conditions during a dynamic gait. 
\end{abstract}


\section{INTRODUCTION}
Robust locomotion strategies and availability of reliable quadruped platforms has pushed the legged locomotion community to add manipulation capabilities to these systems by equipping them with manipulators. Such robots in unstructured environments require reactive strategies to external forces to avoid compromising their balance, e.g for autonomous or tele-operated missions, or to adapt their compliance when interacting with a human, e.g in a collaborative task.

Regarding control strategies for interaction, impedance control \cite{ImpdanceControlMainBook} is a technique widely used on fixed-base robot manipulators and also already implemented on legged robots \cite{976014} \cite{10.1177/0278364915578839} \cite{8246892}. Many are the reasons behind the success of this technique, for example: it allows for indirect force control while establishing a dynamic relationship between the external forces and the motion quantities (positions, velocities and acceleration) of the robot links or end-effector; it does not require any explicit switching strategy between free motion and contact, making it suitable for robot-environment or robot-human interaction tasks; the closed-loop response resembles the dynamics of a second-order system where its response, for motion tracking and under the presence of external forces, can be designed through gains that mimic the effect of spring and damper components; if the information about the external force is available for feedback (from sensors or observer), it is also possible to perform the so called \textit{inertia-shaping} (changing the apparent inertia of links or end-effector) to design the full motion response for the interaction.

\begin{figure}[t!]
     \centering
     \begin{subfigure}
         \centering
         \includegraphics[scale=0.7,width=\columnwidth]{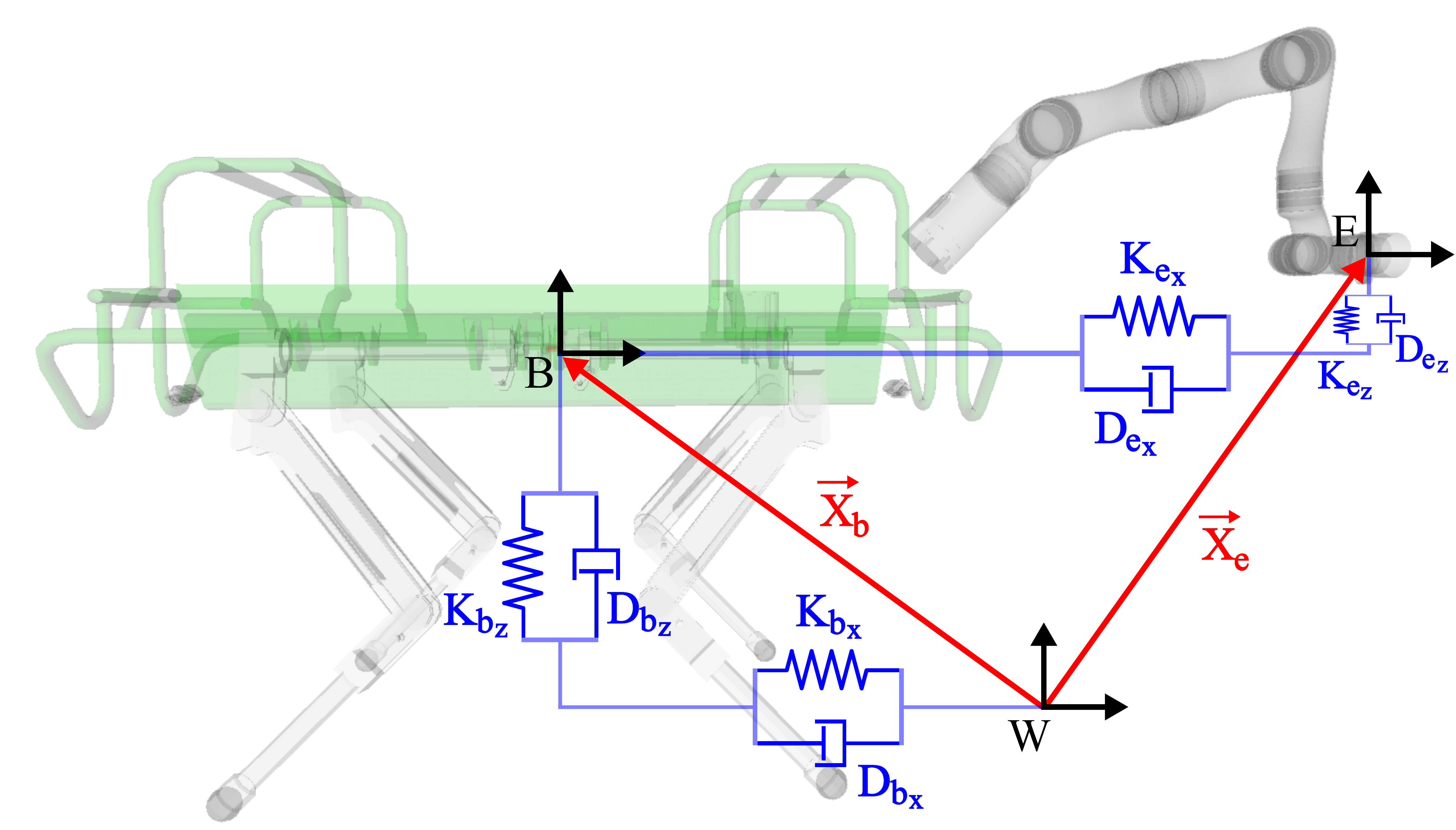}
         \label{fig:quad_plus_arm}
     \end{subfigure}
     \begin{subfigure}
         \centering
         \includegraphics[width=0.8\columnwidth]{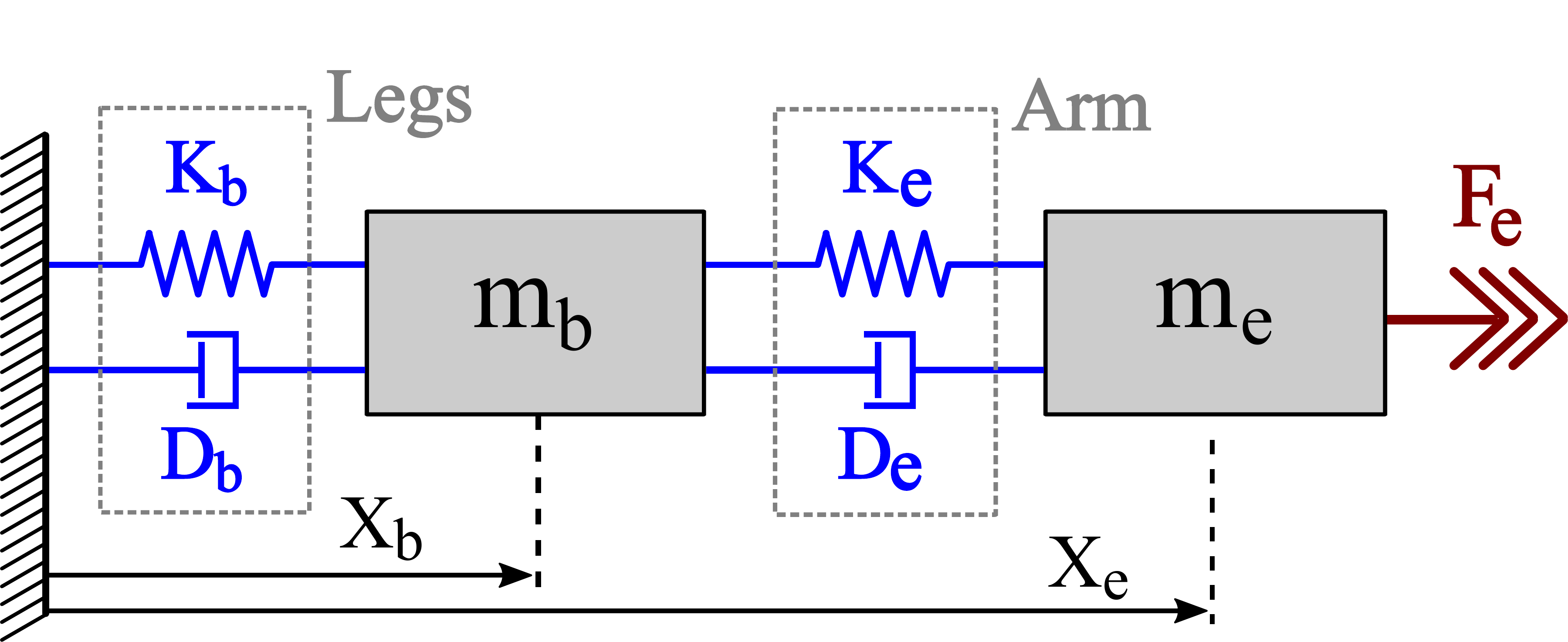}
         \label{fig:simplified_template}
     \end{subfigure}
        \caption{Top: HyQ robot \cite{semini10phdthesis} equipped with a 7-DoF manipulator arm (Kinova Gen3 \cite{kinovaGen3}), its main reference frames (W: inertial, B: robot base, E: arm's end-effector), the robot base and end-effector positions, and the virtual spring and damper impedance components for the sagittal plane. Bottom: Illustration of the simplified template that represents the closed-loop behavior of the whole system for each of the Cartesian dimensions (x, y and z). The gray dashed rectangles highlight the virtual elements that are associated to the legs and the arm.}
        \label{fig:my_label}
\end{figure}

For multi-degree of freedom (DOF) floating-base systems, such as quadrupeds or humanoids, optimization-based techniques in the form of a quadratic program (QP) are prominently adopted for whole-body control (WBC) design \cite{sentis2005synthesis}, \cite{bellicoso2016perception}, \cite{Risiglione}. Such control schemes optimize quadratic control objectives for multiple tasks while handling physical constraints, such as actuation limits, friction cone constraints, kinematic limits, etc. 
To generate motion and maintain balance, previous works on quadrupedal manipulators focused on controlling the center of mass \cite{7487545} \cite{fahmi2019passive} or the torso and the limbs independently  constraining the robot’s Zero-Moment Point to lie inside the support polygon \cite{bellicoso2019alma}.

In the contest of whole-body control, Cartesian impedance control has been synergistically integrated in a QP framework. Results have been shown to regulate impedance of a humanoid upper-body robot \cite{8462877} and for a quadrupedal robot during a static walk \cite{Xin2020AnOL}. An extension \cite{impedanceOnlineAdaptationQuadruped} of the latter work considers variable gains to minimize tracking error without violating a stability criteria. A strong assumption is however made by the authors regarding the Operational space inertia matrix of the torso of the quadruped, considered as constant. 
Changing the desired Cartesian inertia requires to feedback external forces \cite{6385690}. In the state of the art this is limited to fixed-base robot \cite{JIANBIN2009105}. Recently, \cite{DIETRICH2021104875} investigated the effects of shaping the inertia for interaction and tracking. As mentioned by the authors, there are advantages in terms of tracking performance, by getting rid of the coupling terms appearing in the natural joint or Cartesian inertia matrix. Furthermore, the control designer has an additional degree of freedom for pole placement. Downscaling the inertia has also been mentioned \cite{4895692} \cite{dlr123988} to increase the fidelty and user transparency in tele-operation since the user can feel the inertia of the manipulated object.  

In this work, we propose a whole-body controller strategy for our quadrupedal manipulator that integrates a Cartesian impedance controller in a QP problem, allowing to shape the whole compliance of the system, in terms of stiffness, damping and both natural inertias of base and arm. The derived closed-loop system is a double mass-damper-spring system, with all physical parameters that can be tuned accordingly. We perform an extensive analysis of the controller capabilities in rendering impedances, considering inertia shaping for the base as well as for the arm end-effector. Additionally, a discussion regarding the rendering of impedances while trotting, the most common gait used in quadruped locomotion for loco-manipulation tasks, is provided. To the best of our knowledge, this is the first time a control strategy for impedance rendering is assessed considering a dynamic locomotion.

The paper is organized as follows: Section \ref{sec:dynamic_model}
describes the equations of motion for the robotic system composed
of a quadruped robot equipped with a manipulator arm. The linear
double-mass spring damper system, used as target behavior for the
whole-body controller, is detailed in Sec. \ref{sec:template}.
The proposed whole-body controller is explained in details in Sec.
\ref{sec:optimization} and its performance regarding impedance
rendering is assessed in Sec. \ref{sec:results}. Section
\ref{sec:conclusions} closes the paper with conclusions and future
work.


\section{ROBOT MODEL} \label{sec:dynamic_model}
The full rigid body dynamics of a legged manipulator can be described by the set of equations in \eqref{eq:fullDynamicModel}, where $\bm M$ is the inertia matrix, $\bm {\dot{u}}$ the stacked vector of generalized accelerations, $\bm h$ comprises the Coriolis and Centrifugal terms and gravity, $\bm {\tau}$ the actuation torques. 
The stacked vector of generalized accelerations  $\bm {\dot{u}} = [\bm {\ddot{q}}^T_b, \bm {\ddot{q}}^T_l, \bm {\ddot{q}}^T_a]^T \in \mathbb{R}^{6+n_l+n_a}$ denotes the linear and angular accelerations of the base $\bm {\ddot{q}}_b = [\bm {\ddot{x}}^T_b, \bm {\dot{w}}^T_b]^T \in \mathbb{R}^6$ and the rest of the limb joint accelerations.
The subscripts \textit{b}, \textit{l}, \textit{a} and \textit{e} stand for base, legs, arm and arm's end-effector, respectively. $\bm {F}_g \in \mathbb{R}^{3n_c}$ are the ground reaction forces, where $n_c$ denotes the number of contact feet;  $\bm {F}_e \in \mathbb{R}^3$ denotes the external force acting on the arm's end-effector. $\bm {F}_g$ and  $\bm {F}_e$ are mapped respectively to the base through the contact Jacobians $\bm {J}^T_{st}$ and $\bm {J}^T_{e}$.    \\
\begin{equation}
\begin{split}
\underbrace{
\begin{bmatrix}
\bm M_b & \bm {M}_{bl} & \bm {M}_{ba} \\
\bm {M}_{l b} & \bm {M}_{l} & \bm {M}_{la} \\
\bm {M}_{a b} & \bm {M}_{a l} & \bm {M}_{a} 
\end{bmatrix}}_\text{$\bm {M}$}
\underbrace{
\begin{bmatrix}  
\bm {\ddot{q}}_b \\
\bm {\ddot{q}}_l\\
\bm {\ddot{q}}_a
\end{bmatrix}}_\text{$\bm {\dot{u}}$}
&+
\underbrace{
\begin{bmatrix}
\bm {h}_b \\
\bm {h}_l \\
\bm {h}_a \
\end{bmatrix}}_\text{$\bm {h}$}
 = 
\underbrace{
\begin{bmatrix}
\bm {J}^{T}_{st, b} \\
\bm {J}^{T}_{st, l} \\
\bm {0}_{ax3n_c} 
\end{bmatrix}}_\text{$\bm {J}^T_{st}$}\bm {F}_{g} + \\ &+ \underbrace{\begin{bmatrix}
\bm {J}^T_{e,b} \\
\bm {0}_{lx3} \\
\bm {J}^T_{e,a} 
\end{bmatrix}}_\text{$\bm {J}^T_{e}$}\bm {F}_e + \underbrace{\begin{bmatrix}
\bm {0}_{6x1} \\
\bm {\tau}_l \\
\bm {\tau}_a 
\end{bmatrix}}_\text{$\bm {\tau}$} 
\end{split}
\label{eq:fullDynamicModel}
\end{equation}

\section{Simplified Template} \label{sec:template}
The legged manipulator can be seen as a system made up of two masses: one at the base of the quadruped and one at the arm's end-effector. Active impedance control allows to establish virtual elements between the masses and the external environment, through the two chains of actuators: legs and arm, as shown in Fig. \ref{fig:my_label}. By exploiting this technique and connecting the masses as shown in Fig. \ref{fig:my_label}, the interconnected system can be described as
\begin{align}
\begin{split}
\bm M_b\bm {\ddot{x}}_b &= -\bm {K}_b\bm {x}_b - \bm {D}_b \bm {\dot{x}}_b + \bm {K}_e(\bm {x}_e - \bm {x}_b) \\&+ \bm {D}_e(\bm {\dot{x}}_e - \bm {\dot{x}}_b)\\
\bm M_e\bm {\ddot{x}}_e &= \bm {K}_e(\bm {x}_b - \bm {x}_e) + \bm {D}_e(\bm {\bm {\dot{x}}_e - \dot{x}}_b) + \bm {F}_e
\label{eq:SimplifiedTemplateDynamics}
\end{split}
\end{align}
where $\bm {x}_b(t=0)=\bm {x}_{b}^d \in \mathbb{R}^3$, $\bm {x}_e(t=0)=\bm {x}_{e}^d \in \mathbb{R}^3$, both quantities expressed in the inertial frame. $\bm {M}_i \in \mathbb{R}^{3\times3}$, $\bm {D}_i \in \mathbb{R}^{3\times3}$ and $\bm {K}_i \in \mathbb{R}^{3\times3}$, for $i = \{b,e\} $, represents the desired mass, damping and spring terms respectively, defined as diagonal positive matrices, and $\bm F_e \in \mathbb{R}^3$ is the external force acting at the arm.
Alternatively, without any restriction on the initial conditions for the base and the arm's end-effector, considering $\bm {x}_e = \bm {x}_b + \bm {x}_{be}$, we can express \eqref{eq:SimplifiedTemplateDynamics} as
\begin{align}
\begin{split}
\bm M_b\bm {\ddot{x}}_b &= \bm {K}_b( \bm {x}^d_b - \bm {x}_b) + \bm {D}_b(\bm {\dot{x}}^d_b - \bm {\dot{x}}_b) + \bm {K}_e(\bm {x}_{be} - \bm {x}_{be}^d) \\&+ \bm {D}_e(\bm {\dot{x}}_{be} - \bm {\dot{x}}_{be}^d)\\
\bm M_e\bm {\ddot{x}}_e &= \bm {K}_e(\bm {x}_{be}^d - \bm {x}_{be}) + \bm {D}_e( \bm {\dot{x}}_{be}^d - \bm {\dot{x}}_{be}) + \bm {F}_e
\label{eq:SimplifiedTemplateDynamicsWithoutOffset}
\end{split}
\end{align}
In \eqref{eq:SimplifiedTemplateDynamicsWithoutOffset} we make the assumption that the only external interaction force applied to the robot is through the arm. Additionally, we limit our analysis to the translational dynamics, leaving the rotational part for a future work. The obtained equations are linear and the response is set by the impedance parameters. In the next section, we show how to render the behavior described by the simplified template in \eqref{eq:SimplifiedTemplateDynamicsWithoutOffset}.

\section{Optimization} \label{sec:optimization}
\label{sec:bilateral_teleoperator_setup}
In this section we provide the fundamentals for the structure of our control approach. In order to follow the desired template, while satisfying physical and actuation constraints, a QP problem is solved to find the ground reaction forces to track the desired Cartesian acceleration of the base and accelerations for tracking the desired end-effector's motion, see \eqref{eq:SimplifiedTemplateDynamicsWithoutOffset},  as follows:

\begin{align}
\min_{\bm \xi = [\bm {\dot{u}}^T, \bm {F}^T_g]^T} \quad & ||\bm M_b\bm {\ddot{q}}_b - \bm {W}_{b}^{d}||^2_{\bm Q_{b}} + ||\bm {\ddot{q}}_a - \bm {\ddot{q}}_a^d||^2_{\bm Q_{a}} \notag
+ ||\bm  \xi||^2_{\bm Q_r}\\
\textrm{s.t.} \quad & \bm A \bm \xi = \bm b \notag\\ 
&\underline{\bm {d}} < \bm {C} \bm \xi < \overline{\bm d}  
\label{eq: optFormulation}
\end{align}
where the decision variables are: base and limbs accelerations ($\bm {\dot{u}}$) and ground reaction forces ($\bm {F}_g$). The tracking part of the cost function is designed at the acceleration level to mimic the desired template. The equality constraints encode dynamic consistency, no slipping or separation of stance legs, and motion generation for the swing legs. The inequality constraints encode friction constraints, joint kinematic and torque limits. All constraints are stacked in the matrix $\bm {A}^T = [\bm {A}_p^T \; \bm {A}_{st}^T  \; \bm {A}_{sw}^T] $ and $\bm {C}^T = [\bm {C}_{fr}^T \; \bm {C}_{j}^T  \; \bm {C}_{\tau}^T] $.  Details are provided in the following subsections. \\
\subsubsection{Cost} The first term in the cost \eqref{eq: optFormulation} represents the tracking error between the actual $\bm {M}_b \bm {\ddot{q}}_b$ and the desired wrench $\bm W_{b}^d$ wrench on the base. The second term represents the tracking error between the actual $\bm {\ddot{q}}_a$ and the desired $\bm {\ddot{q}}^d_{a}$ acceleration. The third term is to regularize the decision variables.
We define 
\begin{align}
\begin{split}
\bm {F}_{b}^d &= \bm {K}_b(\bm {x}_b^d - \bm {x}_b) + \bm {D}_b(\bm {\dot{x}}^d_b - \bm {\dot{x}}_b) + \bm {K}_e(\bm {x}_{be} - \bm {x}_{be}^d) \\& + \bm {D}_e(\bm {\dot{x}}_{be} - \bm {\dot{x}}_{be}^d) 
\end{split} \\
\bm T_{b}^d &= \bm {D}_r(\bm {w}_{b}^d - \bm {w}_b) + \bm {K}_r\bm {e}_r
\end{align}
where $\bm {F}_{b}^d$ and $\bm T_{b}^d$ are respectively the desired force and moment for the base, i.e ${\bm {W}_{b}^d = [{\bm {F}^d_{b}}^T, {\bm T^d_{b}}^T]}^T$. We define as $\bm {e}_r$ the rotational error, $\bm {D}_r \in \mathbb{R}^{3\times3}$ and $\bm {K}_r \in \mathbb{R}^{3\times3}$ diagonal gain matrices for the derivative and proportional term, respectively.
Let consider the  kinematic relationship between the arm joint velocities and Cartesian velocity expressed in the inertial frame, i.e $\bm {\dot{x}}_e = \bm {\dot{x}}_b + \bm{R}_{wb} \bm {J}_a \bm {\dot{q}}_a$, where $\bm {\dot{x}}_b$ is the measured base velocity in the inertial frame, $\bm{R}_{wb}$ is the rotation matrix from base to inertial frame, $\bm J_a = \frac{\partial \bm {r}_{be}}{\partial \bm {q}_a} \in \mathbb{R}^{6\times n_a}$ expressed in base frame. By deriving the relationship over time, we get 
\begin{equation}
\bm {\ddot{x}}_e = \bm {\ddot{x}}_b + \bm {\dot{R}}_{wb} \bm {J}_a \bm {\dot{q}}_a + \bm{R}_{wb} \bm {\dot{J}}_a \bm {\dot{q}}_a + \bm{R}_{wb} \bm {J}_a \bm {\ddot{q}}_a   
\label{eq: kineamticRelationshipArm}
\end{equation}
By using  \eqref{eq: kineamticRelationshipArm}, the desired joint arm accelerations, $\bm {\ddot{q}}_a^d$, to track the Cartesian acceleration, $\bm {\ddot{x}}_e$ \ref{eq:SimplifiedTemplateDynamics}, are found as
\begin{equation}
\begin{split}
\bm {\ddot{q}}_{a}^d &= \bm {J}_a^{+}\{\bm {R}_{bw} \bm {M}_{e}^{-1}[\bm {K}_e(\bm {x}_{be}^d - \bm {x}_{be}) + \bm {D}_e(\bm {\dot{x}}_{be}^d - \bm {\dot{x}}_{be}) \\&+ \bm {F}_e] -\bm {R}_{bw}\bm {\ddot{x}}_b - \bm {\dot{J}}_a\bm {\dot{q}}_a -\bm {R}_{bw} \bm {\dot{R}}_{wb}\bm {J}_a \bm {\dot{q}}_a \} + \bm {N}_a\bm {\ddot{q}}_{a}^p  
\label{eq: desiredJointAcc}
\end{split}
\end{equation}
where $\bm {\ddot{x}}_b$ is the measured base acceleration, $\bm {J}_a^+$ is the dynamically consistent pseudo-inverse, and $\bm {\ddot{q}}_{a}^p$ is an acceleration defined to keep a desired posture in the null space of the primary element, 
\begin{equation}
\bm {\ddot{q}}_{a}^p = \bm {K}_p(\bm {q}_a - \bm {q}_a^d) - \bm {D}_p \bm {\dot{q}}_a   
\end{equation}
To avoid numerical issues around singular configurations, a damped pseudo-inverse is employed, where the damping term is modulated according to the manipulability of the arm as in \cite{Nakamura86}.
\subsubsection{Physical consistency} To enforce physical consistency constraint between $\bm {F}_g$ and accelerations we enforce both quantities to respect the unactuacted dynamics of the base, i.e.,
\begin{equation}
\begin{aligned}
\bm {A}_p &= 
\begin{bmatrix}
\bm {M}_b & \bm {M}_{bj} & \bm {M}_{ba} & -\bm {J}^T_{st,b}
\end{bmatrix},\\
\bm {b}_p &= -\bm {h}_b + \bm {J}^T_{c,b}\bm {F}_e   
\end{aligned}
\label{eq: pcc}
\end{equation}
Since the ground reaction forces do not appear directly in the cost as a tracking term, this is the constraint used by the optimizer to regularize their values.
\subsubsection{Stance constraint} We enforce zero acceleration for the feet in stance. The stance velocity, $\bm {\dot{x}}_{st}$, is defined as
\begin{equation}
\bm {\dot{x}}_{st} = 
\begin{bmatrix}
\bm {J}_{st,b} & \bm {J}_{st,l}
\end{bmatrix}
\begin{bmatrix}
\bm {\dot{q}}_b \\
\bm {\dot{q}}_l
\end{bmatrix} = 0
\label{ed: stance_constraint}
\end{equation}
which differentiated over time leads to rewrite \eqref{ed: stance_constraint} in matrix form as 
\begin{align}
\bm {A}_{st} &= 
\begin{bmatrix}
\bm {J}_{st,b} & \bm {J}_{st,l} & \bm {0}_{3n_c\times n_a} & \bm {0}_{3n_c\times 3n_c}
\end{bmatrix}\\
\bm {b}_{st} &= -\bm {\dot{J}}_{st,b}\bm {\dot{q}}_b - \bm {\dot{J}}_{st,l}\bm {\dot{q}}_l  
\end{align}

\subsubsection{Swing task} We constrain each swing foot to follow a desired acceleration relative to the base frame. Let denote by $\bm {\ddot{x}}_{sw}^d \in \mathbb{R}^{n_{sw}}$ the stacked vector of desired swing foot accelerations, given the number of swing legs $n_{sw}$. The motion constraint can be defined as
\begin{equation}
\bm {\ddot{x}}_{sw}^d = \bm {J}_{sw}\bm {\ddot{q}}_{sw} + \bm {\dot{J}}_{sw}\bm {\dot{q}}_{sw}     
\end{equation}
rewritten in matrix form as
\begin{equation}
\begin{aligned}
\bm {A}_{sw} &= 
\begin{bmatrix}
\bm {0}_{n_{sw} \times 6} & \bm {J}_{sw} &  \bm {0}_{n_{sw}\times n_a} & \bm {0}_{n_{sw}\times n_{st}}  
\end{bmatrix},\\
\bm {b}_{sw} &= \bm {\ddot{x}}_{sw}^d - \bm {\dot{J}}_{sw}\bm {\dot{q}}_{sw}   
\end{aligned}
\end{equation}
The stacked Jacobian $\bm {J}_{sw}$ maps joint velocities to Cartesian swing foot velocities relative to the base frame. A Cartesian proportional-derivative action is used to set the desired value for $\bm {\ddot{x}}_{sw}^d$, given the desired position and velocities of each swing foot.
\subsubsection{Friction constraints}
We impose contact forces to lie inside the friction cones and their normal component to respect maximum/minimum force threesholds. To incorporate unilaterality on the ground reaction forces an almost zero lower bound is used. We approximate friction cones with square piramids to express them with linear constraints. Hence, friction constraints are expressed as inequalities constraints as
\begin{equation}
\underline{\bm{d}}_{fr} < \vc{C}_{fr}\bm{u} < \overline{\bm{d}}_{fr}, \quad
\bm{C}_{fr} =
\begin{bmatrix}
\bm {0}_{p\times(6+n_l+n_a)} & \bm {F}_{fr}
\label{eq:frictionIneq}
\end{bmatrix}
\end{equation}
with
\begin{equation}
{\bm {F}}_{fr} = \left[
\begin{array}{ccc}
   \bm {F}_0 & \cdots & 0 \\
   \vdots & \ddots & \vdots \\
   0 & \cdots & \bm {F}_c
\end{array}
\right], 
\bm {\underline{d}}_{fr} = \left[
\begin{array}{c}
   \bm {\underline{f}}_0  \\
   \vdots  \\
   \bm {\underline{f}}_c
\end{array}
\right]
,
\overline {{\bm d}}_{fr} = \left[
\begin{array}{c}
   \bm {\overline{f}}_0  \\
   \vdots  \\
   \bm {\overline{f}}_c
\end{array}
\right]
\end{equation}
where $p = 6n_c$ denotes the number of inequalities constraints,
$\bm {F}_{fr} \in {\mathbb{R}}^{p\times3n_c}$ encodes friction cone boundaries for each stance leg, and $\bm {\underline{d}}_r$ and $\bm {\overline{d}}_r$ are lower and uppers bounds, respectively. For further implementation details of the friction constraints we refer to \cite{Focchi2017HighslopeTL}.
\subsubsection{Joint kinematic limits} To avoid violation of joint kinematic limits, limb accelerations are bounded through the following inequality constraint
\begin{equation}
\underline{\bm {{d}}}_j < \bm {C}_j\bm {u} < \overline{\bm {{d}}}_j
\end{equation}
\begin{equation}
\bm {C}_j = 
\begin{bmatrix}
\bm {0}_{(n_l+n_a)\times 6} & \bm {I}_{(n_l+n_a)\times(n_l+n_a)} & \bm {0}_{(n_l+n_a)\times 3n_c}
\end{bmatrix}
\end{equation}
\begin{equation}
\underline{\bm {d}}_j = \bm {\ddot{{q}}}_{j_{\min}}(\bm {q}_j), 
\; \overline{\bm d}_j = \bm {\ddot{{q}}}_{j_{\max}}(\bm {q}_j)
\end{equation}
As proposed in \cite{fahmi2019passive}, the bounds, $\bm {\underline{d}}_j$ and $\bm {\overline{d}}_j$, are re-computed at each time step, according to the joint configurations, and set to reach zero accelerations in a time interval $\Delta t$
\begin{equation}
\ddot{\vc{q}}_{j_{\min,\max}} = -\frac{2}{\Delta t^2}(\vc{q}_{j_{\min,\max}} -
\vc{q}_j - \Delta t\, \dot{\vc{q}}_j).
\label{eq:accellBounds}
\end{equation} 

\subsubsection{Torque limits} We enforce actuation limits from the decision variables by exploiting the actuated part of the dynamics \eqref{eq:fullDynamicModel} and the control law \eqref{eq:inverseDynamics} as
\begin{equation}
\bm {\underline{d}}_{\tau} < \bm {C}_{\tau}\bm {u} < \bm {\overline{d}}_{\tau}
\end{equation}
\begin{equation}
\underline{\bm {d}}_{\tau} = \begin{bmatrix}
-\bm {h}_l  + \bm {\tau}_{l,min}(\bm {q}_l) \\
-\bm {h}_a + \bm {J}_{c,a}^T\bm {F}_e + \bm {\tau}_{a,min}
\end{bmatrix}
\end{equation}
\begin{equation}
\overline{\bm {d}}_{\tau} = \begin{bmatrix}
-\bm {h}_l + \bm {\tau}_{l,max}(\bm {q}_l) \\
-\bm {h}_a + \bm {J}_{c,a}^T\bm {F}_e + \bm {\tau}_{a,max}
\end{bmatrix} 
\end{equation}
\begin{equation}
\bm {C}_r = \begin{bmatrix}
\bm {M}_{lb} & \bm {M}_l & \bm {M}_{la} & -\bm{J}^T_{st,l}\\
\bm {M}_{ab} & \bm {M}_{al} & \bm {M}_{a} & -\bm{0}_{ax3n_c}
\end{bmatrix}    
\end{equation}

\subsubsection{Torque computation}
We map the optimal solution $\bm {\xi} = [\bm {\dot{u}}^T, \bm {F}^T_g]^T$ obtained by solving the QP into desired torques $\bm {\tau} = [\bm {\tau}^T_l, \bm {\tau}^T_a]^T$ for limbs (legs and arm) using the full dynamics in equation \eqref{eq:fullDynamicModel} as
\begin{equation}
\begin{split}
 \begin{bmatrix}
\bm {\tau}_l\\
\bm {\tau}_a
\end{bmatrix} &=
\begin{bmatrix}
\bm {M}_{lb} & \bm {M}_l & \bm {M}_{la}\\
\bm {M}_{ab} & \bm {M}_{al} & \bm {M}_a
\end{bmatrix}
\begin{bmatrix}
\bm {\ddot{q}}_b \\
\bm {\ddot{q}}_l \\
\bm {\ddot{q}}_a 
\end{bmatrix} +
\begin{bmatrix}
\bm {h}_{l}\\
\bm {h}_{a}
\end{bmatrix} -
\begin{bmatrix}
\bm {J}^T_{st,l}\\
\bm {J}^T_{st,a}
\end{bmatrix}\bm {F}_g \\&-
\begin{bmatrix}
\bm {0}_{lx3n_c}\\
\bm {J}^T_{c,a}
\end{bmatrix}\bm{F}_e
\end{split}
\label{eq:inverseDynamics}
\end{equation}

\section{IMPLEMENTATION DETAILS}
Hard constraints can lead to infeasibility if conflicting with each other. This can occur in our proposed approach since all the constraints have the same priority. A commonly used solution is to relax constraints through the use of slack variables. However, we cannot relax torque and joint constraints because they cannot be violated. Same applies for the physical consistency, friction and stance constraints which results necessary to find ground reaction forces that avoid slippage and mantain balance. On the other hand, swing tasks cannot always be fulfilled due to the limited motion of the legs. Slack variables, $\bm s$, are used here to relax the swing constraints, as in \cite{fahmi2019passive}, i.e
\begin{equation}
-\bm{s} \le \bm {J}_{sw}\bm {\ddot{q}}_{sw} + \bm {\dot{J}}_{sw}\bm {\dot{q}}_{sw} - \bm {\ddot{x}}_{sw}^d \le \bm{s}   
\end{equation}
\begin{equation}
\bm{s} \ge 0    
\label{eq: positivitySlackVariables}
\end{equation}
where \eqref{eq: positivitySlackVariables} ensures the slack variables to be non-negative.
The decision variables are augmented as $\bm {\xi} = [\bm {\dot{u}}^T, \bm {F}^T_g, \bm {s}^T]^T$ and the slack variables are penalized as well in the regularization term in \eqref{eq: optFormulation}.


\section{Results} \label{sec:results}
We validated our approach with a set of simulations where
external forces are applied at the arm's end-effector and
the impedance rendering is assessed considering two
scenarios: 1) robot standing still with all legs in contact
with the ground; and 2) robot in place executing a trotting
gait. With these two scenarios, the control capabilities in
rendering the base and end-effector impedances are assessed 
under different contact conditions. The quadrupedal manipulator
is referred to as the \textit{simulated system} and the simplified
template, the double-mass spring system described in (\ref{eq:SimplifiedTemplateDynamicsWithoutOffset}) as the
\textit{reference system}. The simulation is run using Gazebo
as physics simulator. The dynamic terms and Jacobians that compose
the equations of motion, described in (\ref{eq:fullDynamicModel}),
used to devise the whole-body controller are computed using
\textit{RobCoGen} \cite{frigerio:2016:robcogen}. For the sake
of space, this section presents the base and end-effector
interaction responses for only the X and Y Cartesian dimensions
(longitudinal and lateral directions), since these are the most
relevant ones for the locomotion stability in legged systems.
For the sake of clarity, in the plots to come the positions of base and end-effector are shown 
as displacements relative to their respective desired (initial) position. All the simulations discussed in the following sections are also included in the accompanying video 
\footnote{The accompanying video is also available at the following YouTube link: \\\href{https://www.youtube.com/watch?v=yVBqBtA2BM0}{https://www.youtube.com/watch?v=yVBqBtA2BM0}}.

\subsection{Inertia shaping}
The joint-space inertia matrix, as well as its corresponding
Cartesian inertia matrix reflected at the end-effector, are non-diagonal matrices whose elements depend on the posture of the robot
(i.e., on the joints positions). When impedance control is performed
without force feedback, the inertia rendered during interaction
tasks cannot be modified and the closed-loop dynamics for every
Cartesian dimension cannot be completely decoupled. To be able to
extend the assessment of the proposed whole-body controller, in
this paper we assume that the interaction forces at the end-effector
can be measured. Therefore, along the next subsections, we considered
desired inertias defined as positive diagonal matrices and we refer
to as \textit{nominal} values the mean values of the diagonal terms in
$\bm {M}_e$ and $\bm {M}_b$ assumed along their allowed workspace.
The response of each subsystem is tuned separately to have a critically damped response, i.e. $d_i^2 = 4m_ik_i$
where $m_i$, $d_i$ and $k_i$ are respectively the mass, damping
and stiffness coefficients.

\subsection{Inertia shaping on the robot base}
\begin{figure}
\centering
\hspace*{-0.1\linewidth}
\vspace{-0.1cm}
\begin{tikzpicture}
\node[inner sep=0pt] (plotTrajectoryPositionMode) at (-0.7,0)
    {\includegraphics[scale=1,width=1.0\columnwidth]{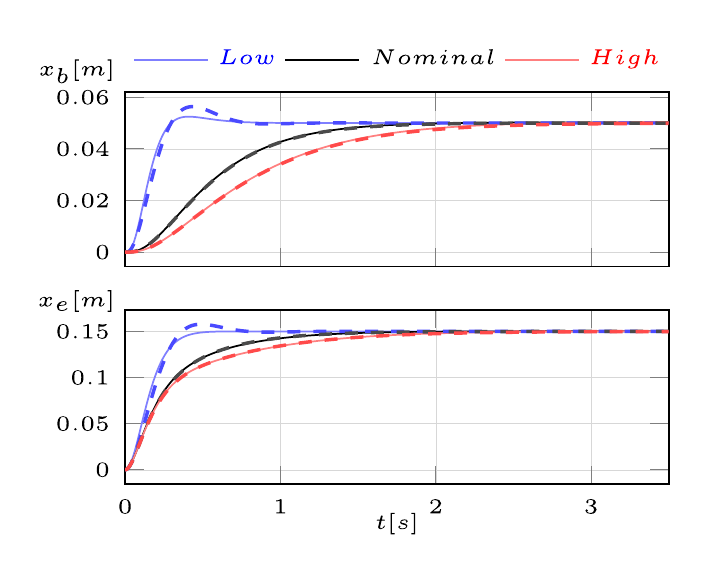}};
\end{tikzpicture}
\caption{Motion response along the longitudinal direction for the robot base (top figure) and end-effector (bottom figure) for different values of base apparent inertia (blue: low mass, black: nominal mass, red: high mass), when a step force of 50 N is applied. The reference system is highlighted with dashed lines, while the simulated system  with solid lines.}
\label{fig: experiment1_step.}
\vspace{-0.0cm}
\end{figure}

In the first set of simulations, we show the effects of performing
inertia shaping on the base considering different mass values.
The inertia of the base is decreased up to the same value of the
end-effector inertia and increased of 200\% of its nominal value. Hence,
three apparent inertias are considered for the base: \textit{low}) base weighting 4 kg; \textit{nominal}) base weighting 92 kg; and \textit{high}) base weighting 184 kg. For the three cases the stiffness is fixed to $K_{b_x}=1000$ N/m and the damping coefficients are calculated as the subsystems were completely decoupled (i.e., two independent second-order systems), resulting in the following values: for low inertia, $D_{b_x}=126.5$ Ns/m; for nominal inertia, $D_{b_x}=610$ Ns/m; and for high inertia $D_{b_x}=858$ Ns/m. For the arm impedance,
the following values are considered: $M_{e_x}=4$ kg, $D_{e_x}=90$ Ns/m,
and $K_{e_x}=500$ N/m.
As shown in Fig. \ref{fig: experiment1_step.}, lowering and
increasing the mass for fixed stiffness gives out, respectively,
slower and faster responses. This is something expected
because the poles of the two subsystem are indeed dependent on
the impedance gains. Even without having a closed-loop solution
that links the impedance gains to the poles of the simplified
template, this suggest us that each subsystem contributes to the
poles of the full simplified template with a real part defined by
a ratio close to $\frac{d_i}{m_i}$. While for the base, where the
time constant is affected by the choice of the mass, the impact
is less noticeable for the arm. Therefore, shaping the inertia
of the base can be used in all the scenarios where locomotion
cannot be sacrificed by forces acting through the arm during
manipulation. It is moreover important to notice that the choice
of the mass does not have any impact on the steady-state value,
being the steady-state only related to the stiffness.

\begin{figure}
\centering
\begin{tikzpicture}
\node[inner sep=0pt] (plotTrajectoryPositionMode) at (-0.7,0)
    {\includegraphics[scale=1.,width=\columnwidth]{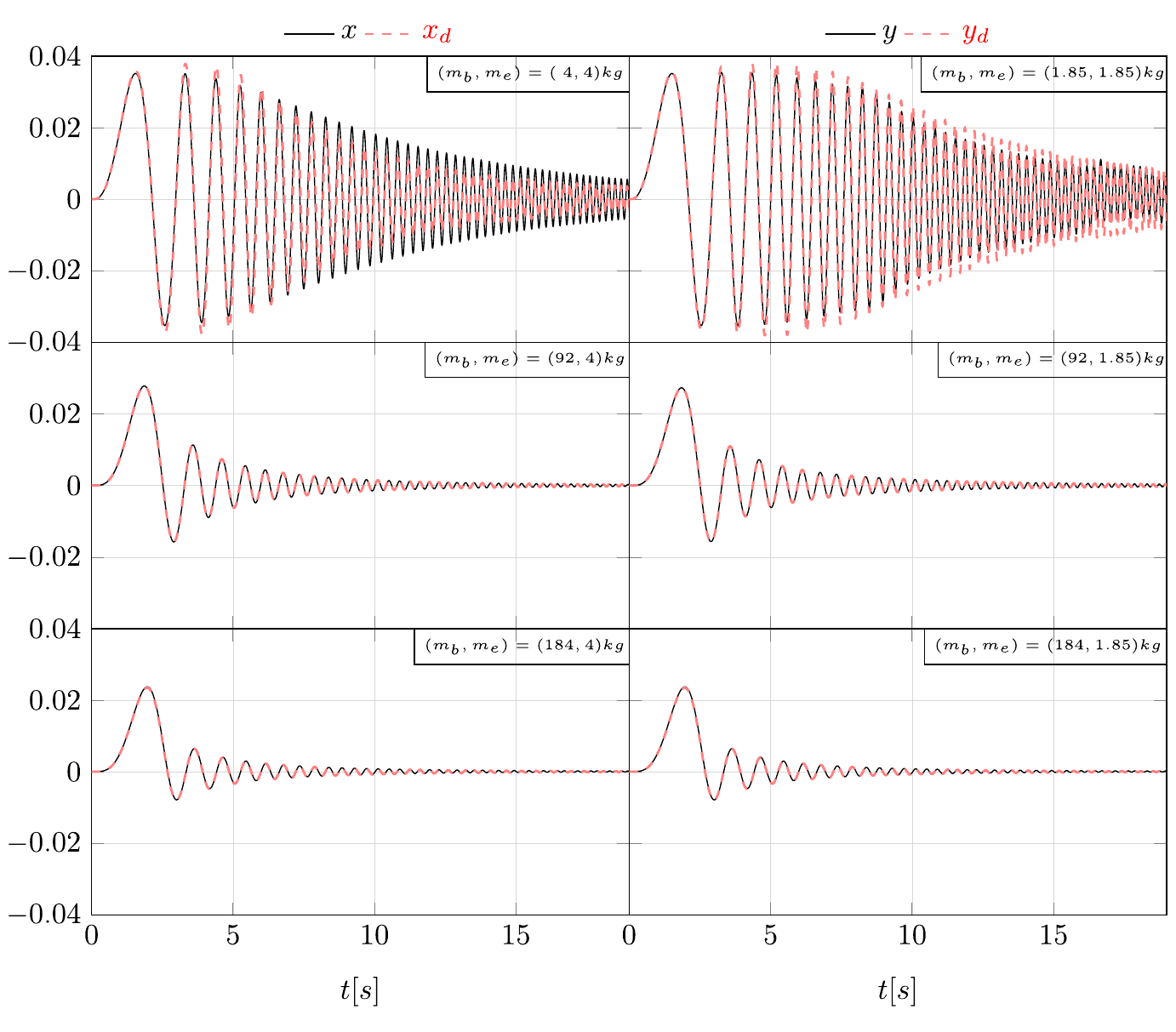}};
\end{tikzpicture}
\caption{Comparison between the response of the base in the reference system (red dashed line) and the simulated system (solid black line) along x and y, respectively on left and right, axis for a chirp signal with frequencies ranging from [0, 5] Hz.}
\label{fig: experimentI_chirp_base.}
\vspace{-0.2cm}
\end{figure}

\begin{figure}[ht!]
\centering
\begin{tikzpicture}
\node[inner sep=0pt] (plotTrajectoryPositionMode) at (-0.7,0)
    {\includegraphics[scale=1.,width=\columnwidth]{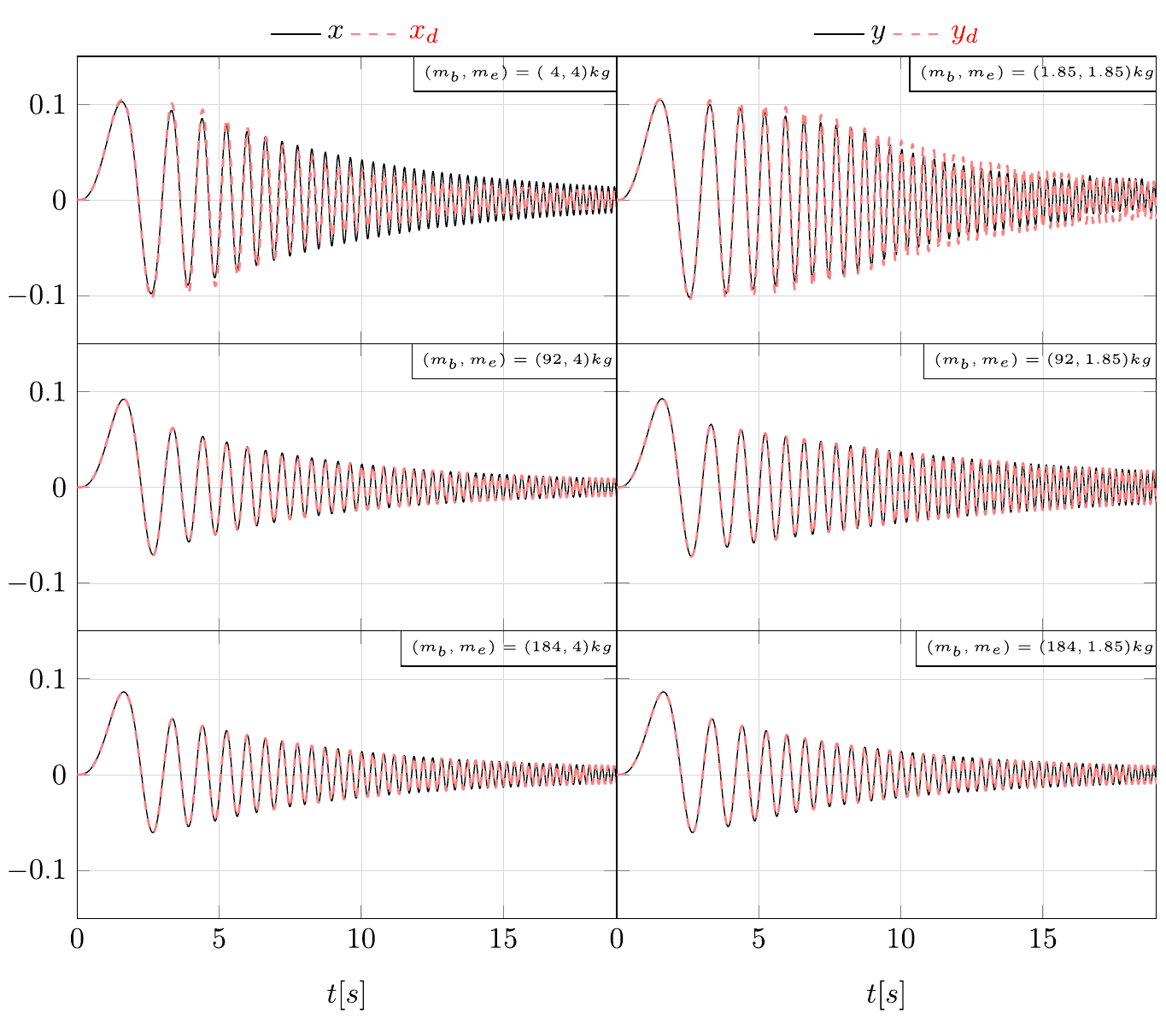}};
\end{tikzpicture}
\caption{Comparison between the response of the end-effector in the reference system (red dashed line) and the simulated system (solid black line) along x and y, respectively on left and right, axis for a chirp signal with frequencies ranging from [0, 5] Hz.}
\label{fig: experimentI_chirp_arm.}
\vspace{-0.2cm}
\end{figure}

To highlight the differences in the transient response according to the
inertia value, we applied a chirp signal with a frequency ranging
from 0 to 5 Hz along $x$ and $y$. Along the two dimensions, the following values for base and arm stiffnesses are considered: $K_{b}=1000$ N/m, $K_{e}=500$ N/m. As shown in Fig. \ref{fig: experimentI_chirp_base.}, we can notice how the base oscillations
decay faster to 0 by increasing the mass of the base. This is
explained by the poles moving to the origin increasing the mass
and therefore lowering the cut off frequency. Analogously to the
step response, the arm is not much affected by the inertia shaping
of the base for nominal and high masses, as seen in Fig. \ref{fig: experimentI_chirp_arm.}.

\subsection{Inertia shaping on the arm}
\begin{figure}[t]
\centering
\hspace*{-0.05\linewidth}
\begin{tikzpicture}
\node[inner sep=0pt] (plotTrajectoryPositionMode) at (-0.7,0)
    {\includegraphics[scale=1,width=1.0\columnwidth]{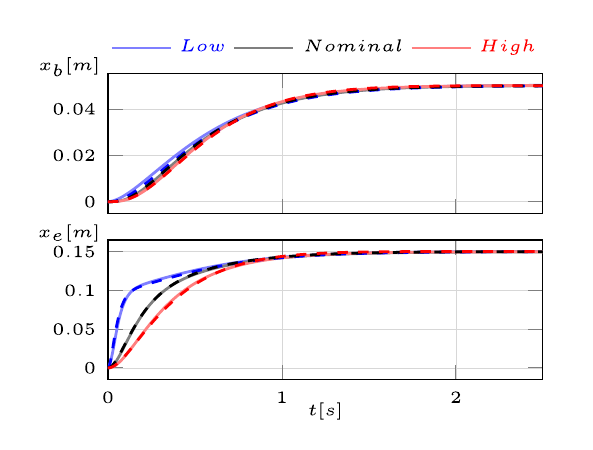}};
\end{tikzpicture}
\caption{Motion response along the longitudinal direction for the robot base (top figure) and end-effector (bottom figure) for different values of end-effector apparent inertia when a step force of 50 N is applied. The reference system is highlighted with dashed lines, while  the simulated system  with solid lines.}
\label{fig:armInertiaShaping}
\vspace{-0.4cm}
\end{figure}

In this section, we show the control capabilities in impedance rendering
when the apparent inertia of the arm is modified. The impedance of 
the arm is varied considering three impedance settings designed for a critically damped response. The base impedance, instead, remains constant and tuned to result in a critical response as well. For the sake of space, only the impedance response for the longitudinal direction is presented. Three apparent inertias are considered for the end-effector: \textit{low}) arm weighting 0.4 kg; \textit{nominal}) arm weighting 4 kg; and \textit{high}) arm weighting 10 kg. For the three cases the stiffness is fixed in $K_{e_x}=500$ N/m and the damping coefficients are calculated as the subsystems were completely decoupled (i.e., two independent second-order systems), resulting in the following values: for low inertia, $D_{e_x}=29$ Ns/m; for nominal inertia, $D_{e_x}=90$ Ns/m; and for high inertia $D_{e_x}=142$ Ns/m. For the base impedance,
the following values are considered: $M_{b_x}=92$ kg, $D_{b_x}=610$  Ns/m,
and $K_{b_x}=1000$ N/m. The system response for the three impedance configurations is showed in Fig. \ref{fig:armInertiaShaping}.

Two observations are important to highlight from Fig. \ref{fig:armInertiaShaping}.
First, the impedance rendering for the arm presents negligible error for the three set of parameters while for the base the behavior presents small deviations from the reference at the beginning of the transient response (where base accelerations are higher). The second observation regards the characteristic response of the double-mass spring damper system that is
clearly manifested dictating the system behavior in the second-half of the transient response for the arm. I.e., being a four-order system in closed-loop, even
if the arm impedance is set to be fast, the convergence of the response is
kept governed by the slow poles associated to the base impedance.

\subsection{Impedance rendering during dynamic gait}
In this section we assess the impedance rendering when the
robot is performing a walking trot. This gait has been
the most used for the research and potential industrial
applications involving loco-manipulation, e.g. as found in
\cite{bellicoso2019alma}, \cite{sleiman2020MPCplanner},
and \cite{9561835}.
However, for the best of the authors knowledge, a study on
how the trotting gait affects an impedance rendering is still
missing in the literature.

A walking trot gait is characterized by the alternated motion
of diagonal leg pairs (left-front leg together with the
right-hind and the right-front together with the left-hind)
with a step duty factor equal or higher than 0.5 (the duty
factor is the ratio between the time a leg is in stance over
the entire step period).

For the robot motion generation and control we exploit the
structure of the Reactive Control Framework \cite{barasuol13icra},
where the trunk controller is substituted by the proposed
whole-body controller. Each foothold (desired touch-down position)
is computed as a function of the robot center-of-mass and the
base velocity given with respected to the horizontal frame
(see \cite{barasuol13icra}).

In this simulation scenario, the impedance rendering is evaluated
for different set of impedance parameters and also according to gait
parameters (step frequency $f_s$ and duty factor $d_f$). As set of impedance
parameters, we consider the two ones corresponding to the fastest
and slowest responses presented in Fig. \ref{fig:armInertiaShaping}.
The renderization accuracy of each impedance set is analysed for
four combinations of gait parameters GP \{$d_f$, $f_s$\}: GP1 = \{0.55, 1.4\},
GP2 = \{0.65, 1.4\}, GP3 = \{0.55, 1.8\} and GP4 = \{0.65, 1.8\}.

\begin{figure}
\centering
\begin{tikzpicture}
\node[inner sep=0pt] (plotTrajectoryPositionMode) at (-0.7,0)
    {\includegraphics[scale=1.0,width=\columnwidth]{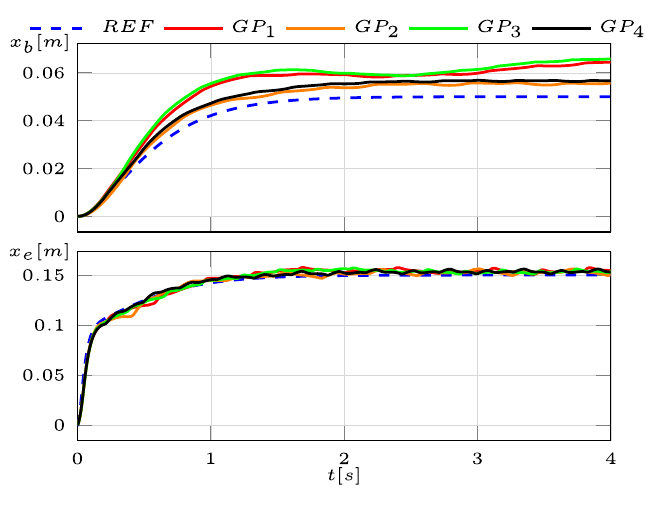}};
\end{tikzpicture}
\caption{Motion response along the longitudinal direction for the
robot base (top figure) and end-effector (bottom figure) for lower
end-effector apparent inertia and four combinations of gait
parameters (GPs).}
\label{fig:trottingLowMass}
\vspace{-0.2cm}
\end{figure}

\begin{figure}
\centering
\begin{tikzpicture}
\node[inner sep=0pt] (plotTrajectoryPositionMode) at (-0.7,0)
    {\includegraphics[scale=1.0,width=\columnwidth]{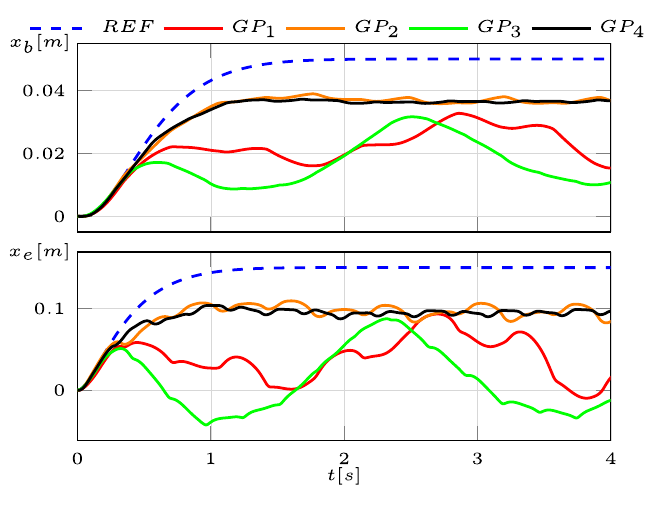}};
\end{tikzpicture}
\caption{Motion response along the longitudinal direction for the
robot base (top figure) and end-effector (bottom figure) for higher
end-effector apparent inertia and four combinations of gait
parameters (GPs).}
\label{fig:trottingHighMass}
\vspace{-0.2cm}
\end{figure}

For the sake of space, the motion response of the base and the
end-effector are only shown for the longitudinal direction ($X$
dimension). Fig. \ref{fig:trottingLowMass} and Fig.
\ref{fig:trottingHighMass} present the results for the impedance
set corresponding, respectively, to the low and high end-effector
apparent inertial.

The most prominent differences appear when comparing the general
behavior of the response for lower and higher end-effector apparent
inertias. The impedance rendering shows to be significantly more
sensitive to trotting gaits for higher apparent inertias. With
respect to gait parameters, the tracking error appears to be
more sensitive to the duty factor rather than the step frequency.
As a trend seen from the plots, tunings with larger duty factors and
lower step frequencies benefits the impedance rendering.

\begin{figure}
\centering
\begin{tikzpicture}
\node[inner sep=0pt] (plotTrajectoryPositionMode) at (-0.7,0)
    {\includegraphics[scale=1.0,width=\columnwidth]{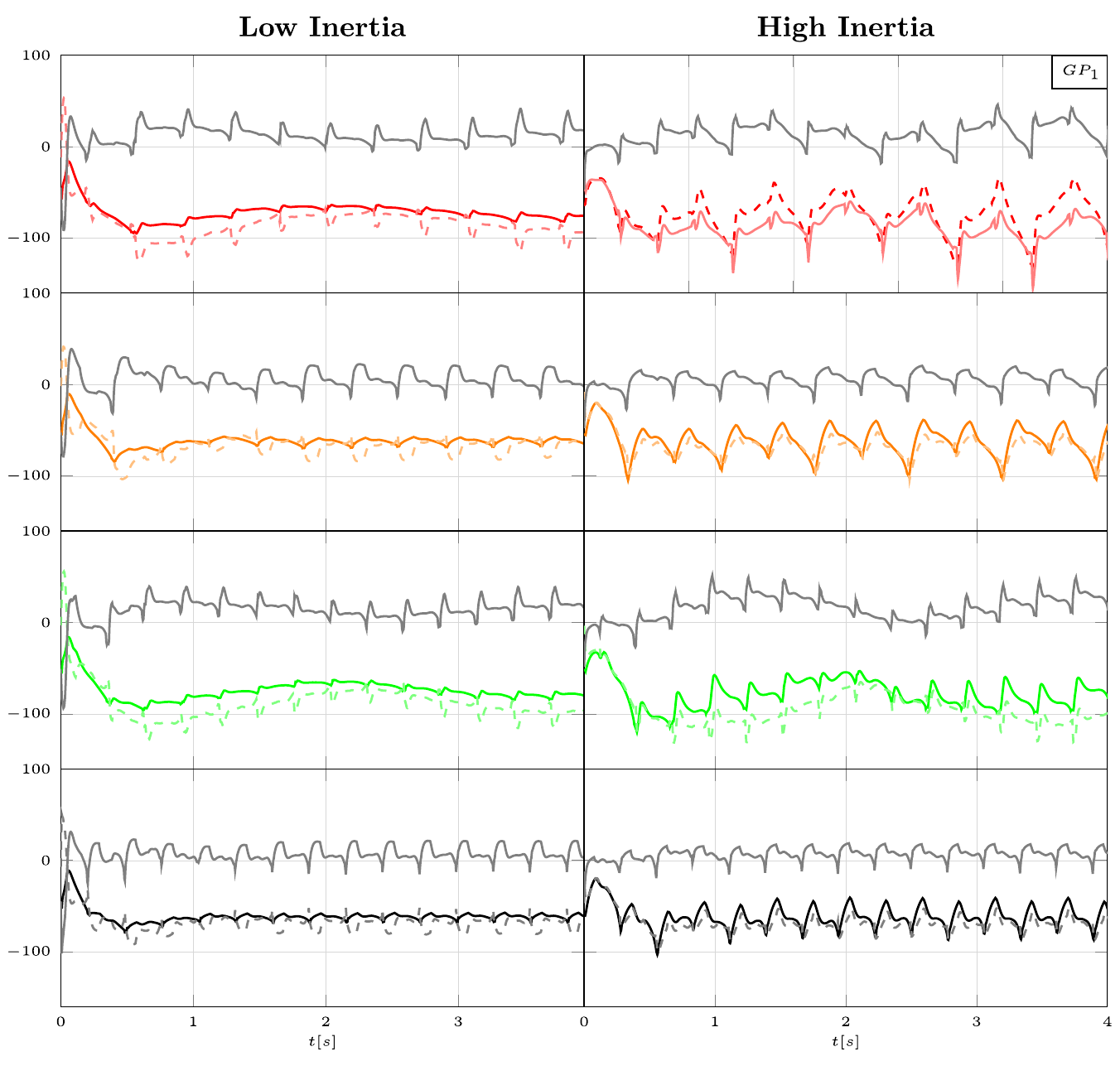}};
\end{tikzpicture}
\caption{Desired (solid line) and applied (dashed line) longitudinal force on the base, and their respective error (gray line) for different gait parameters and end-effector apparent inertia. }
\label{fig:trottingWrenches}
\end{figure}

The interaction between all these four elements (impedance, gait,
step frequency and duty factor) happen with the following
relationships: 1) when there are only two legs on the ground, the
system becomes underactuated and it is not possible to produce
every desired wrench on the base; 2) the higher the apparent inertia,
damper or stiffness the controller seeks to render, the higher the
magnitude of the ground reaction forces requested for that; 3) the higher
the duty factor, the higher is the period in which the system is
fully actuated, allowing the controller to recover tracking; and
4) higher step frequencies require higher leg joint accelerations,
that in turns, produce inertial forces that disturb the base and
have to be counteract by the legs in contact to render the
impedance. All this interaction can also be verified in Fig.
\ref{fig:trottingWrenches}, where the differences between the
desired and achieved wrench components along the $x$ dimension
are shown for all the eight impedance and gait parameters
combinations.

\section{Conclusions} \label{sec:conclusions}

In this paper we introduced a whole-body controller, for
quadruped robots equipped with a manipulator arm, that allows
for controlling the robot base and end-effector as two
subsystem. Together, these two subsystems resemble the dynamics
of a linear double-mass spring damper system. Enforcing
such linear template model on a complex non-linear system
has the advantage of simplifying the control design
concerned about the interaction forces during loco-manipulation.

Extensive analysis of the controller capabilities in
rendering impedances was presented by considering also the
inertia shaping of the base and the arm end-effector. 

We presented, for the first time, a detailed analysis on
how a trotting gait affects the accuracy in rendering
the desired impedance and how gait parameters also
affect the impedance tracking. Results showed that increasing the duty factor has much more impact in reducing the rendering errors than increasing the stepping frequency.

As future work, we aim to expand the impedance
rendering analysis, also considering experimental results and
the implementation of disturbance observers to estimate
external forces on the end-effector.

\bibliographystyle{./bibtex/IEEEtran} 
\bibliography{root}
\addtolength{\textheight}{-12cm}   
\end{document}